\documentclass[letterpaper]{article} % DO NOT CHANGE THIS
\usepackage{aaai2026}  % DO NOT CHANGE THIS
\PassOptionsToPackage{hyphens}{url}
\usepackage{booktabs}
\usepackage{amssymb}
\usepackage{siunitx}
\usepackage{times}  % DO NOT CHANGE THIS
\usepackage{helvet}  % DO NOT CHANGE THIS
\usepackage{courier}  % DO NOT CHANGE THIS
\usepackage[hyphens]{url}  % DO NOT CHANGE THIS
\usepackage{amsmath}
\usepackage{graphicx} % DO NOT CHANGE THIS
\usepackage{natbib}  % DO NOT CHANGE THIS AND DO NOT ADD ANY OPTIONS TO IT
\usepackage{caption} % DO NOT CHANGE THIS AND DO NOT ADD ANY OPTIONS TO IT
\usepackage{algorithm}
\usepackage{algorithmic}

\usepackage{newfloat}
\usepackage{listings}
\DeclareCaptionStyle{ruled}{labelfont=normalfont,labelsep=colon,strut=off} % DO NOT CHANGE THIS
\floatstyle{ruled}
\newfloat{listing}{tb}{lst}{}
\floatname{listing}{Listing}
\title{Lightweight Inference-Time Personalization for Frozen Knowledge Graph Embeddings}
\author{
  Çerağ Oğuztüzün\textsuperscript{\rm 1}\equalcontrib,
  Ozan Oğuztüzün\textsuperscript{\rm 2}\equalcontrib
}
\affiliations{
  \textsuperscript{\rm 1}Department of Computer and Data Sciences, Case Western Reserve University\\
  \textsuperscript{\rm 2}Department of Mechanical Engineering, Case Western Reserve University\\
  Cleveland, OH, USA\\
  cxo147@case.edu, oxo130@case.edu
}

\begin{document}

\maketitle

\begin{abstract}
  Foundation models for knowledge graphs (KGs) achieve strong cohort-level performance in link prediction, yet fail to capture individual user preferences; a key disconnect between general relational reasoning and personalized ranking. We propose GatedBias, a lightweight inference-time personalization framework that adapts frozen KG embeddings to individual user contexts without retraining or compromising global accuracy. Our approach introduces structure-gated adaptation: profile-specific features combine with graph-derived binary gates to produce interpretable, per-entity biases, requiring only ${\sim}300$ trainable parameters. We evaluate GatedBias on two benchmark datasets (Amazon-Book and Last-FM), demonstrating statistically significant improvements in alignment metrics while preserving cohort performance. Counterfactual perturbation experiments validate causal responsiveness; entities benefiting from specific preference signals show 6--30$\times$ greater rank improvements when those signals are boosted. These results show that personalized adaptation of foundation models can be both parameter-efficient and causally verifiable, bridging general knowledge representations with individual user needs.
\end{abstract}


\section{Introduction}

Foundation models for knowledge graphs (KGs) and relational data achieve remarkable \emph{cohort-level} accuracy in link prediction. Yet, practical deployments increasingly demand \textit{profile-conditioned behavior}: the same candidate entity should rank differently for different users, contexts, or patients. Existing approaches typically fine-tune the backbone model or attach trainable adapters to handle each profile, but such strategies introduce substantial overhead (such as new gradient updates, hyperparameter tuning, and retraining cost) while risking degradation of the original cohort performance.

We propose a lightweight, inference-time personalization framework for frozen KG embeddings that adds interpretable, structure-aware biases instead of modifying the backbone. The key idea is \textit{gate personalization through graph structure}: entity attributes extracted from the training KG serve as binary gates, while profile features act as conditioning signals over these gates. This separation allows us to personalize candidate rankings with only a few trainable parameters, without any gradient flow through the backbone model.

%This design preserves the advantages of foundation KG embeddings (such as robust generalization and calibration) while introducing minimal, interpretable adjustments that adapt to individual profiles. %Our empirical results across multiple datasets show consistent personalization gains measured by Alignment@k and Counterfactual Responsiveness (CR), without loss in mean reciprocal rank (MRR) or Hits@k. 

\textbf{Contributions.}
\begin{enumerate}
    \item We introduce a \textbf{post-hoc personalization mechanism} that operates entirely at inference time on frozen KG embeddings, requiring no backbone updates.
    \item We propose \textbf{structure-gated adaptation}, an interpretable way to condition candidate rankings on profile-specific features via graph-derived gates.
    \item We propose \textbf{evaluation metrics for personalization} Alignment@k and Counterfactual Responsiveness to quantify alignment and causal responsiveness of personalized predictions.
\end{enumerate}

%\vspace{0.5em}
Our findings suggest that simple, structure-aware bias adaptation can serve as a general plug-in for personalized ranking in any pretrained KG or relational foundation model, bridging the gap between cohort-level embeddings and individualized predictions.

\section{Related Work}

\textbf{Joint KG–User Learning.}
RippleNet~\cite{wang2018ripplenet}, KPRN~\cite{wang2019explainable}, and KGAT~\cite{KGAT19} jointly embed users and entities, propagating preferences over multi-hop neighborhoods. Effective but deployment-heavy: they retrain per population and alter the embedding space. Our method is post-hoc and keeps the backbone frozen.

\textbf{Parameter-Efficient KGE Adaptation.}
Adapter/LoRA-style methods (e.g., IncLoRA, FastKGE~\cite{liu2024fast}) insert trainable modules to specialize embeddings, yet still require backprop through the backbone and tuning. We train small, independent bias heads without touching backbone gradients.

\textbf{Calibration \& Post-hoc Adjustment.}
Platt scaling and isotonic regression~\cite{tabacof2019probability,safavi2020evaluating,nascimento2024n} improve confidence calibration but are profile-agnostic. Our structure-gated bias is likewise post-hoc but yields profile-specific adjustments.

\textbf{Distinctions.}
Frozen embeddings; no backbone gradients; per-profile (not global) score shifts.

\section{Methods}

\subsection{Problem Setup and Notation}

Let $\mathcal{G} = (\mathcal{E}, \mathcal{R}, \mathcal{T})$ be a knowledge graph (KG) with entities $\mathcal{E}$, relation types $\mathcal{R}$, and triples $\mathcal{T} \subseteq \mathcal{E} \times \mathcal{R} \times \mathcal{E}$.  
Given a query $(h,r,?)$ with head $h \in \mathcal{E}$ and relation $r \in \mathcal{R}$, link prediction ranks candidate tails $t \in \mathcal{E}$ by a pretrained \emph{frozen scorer} $s_\theta(h,r,t)$ (e.g., DistMult, ComplEx, or RotatE).  
The backbone parameters $\theta$ remain fixed during personalization.

Our goal is \textbf{profile-conditioned re-ranking}: for a given profile $p$, the final score is
\[
s'(h,r,t) = s_\theta(h,r,t) + b_p(t),
\]
where $b_p(t)$ is a \emph{tail-specific bias} that depends on both the structure of $t$ in the training graph and the profile’s features.  
Because adding a constant bias to all tails does not change their rank, $b_p(\cdot)$ must vary with $t$ to meaningfully alter rankings.  
Our method makes this dependence explicit and interpretable through structure-gated personalization.

\subsection{Structure-Gated Bias Construction}

We partition relations into $K$ semantic groups $\{\mathcal{R}_k\}_{k=1}^{K}$ (typically $K=2$ for interpretability).  
For each group $k$, we define an \emph{attribute universe}
\[
\mathcal{U}_k = \{\,u \in \mathcal{E} : \exists (u,r,t) \in \mathcal{T}_{\text{train}},\, r \in \mathcal{R}_k \,\}.
\]
Each candidate tail $t$ is associated with a binary \emph{gate vector} 
\[
g_k(t)[i] = \mathbb{1}\!\left\{\, u_i \;\xrightarrow{\;r \in \mathcal{R}_k\;}\; t \text{ in train} \,\right\},
\]

indicating which group-$k$ attributes connect to $t$ in the training graph.  
Stacking these yields gate matrices $G_k \in \{0,1\}^{|\mathcal{E}| \times |\mathcal{U}_k|}$, ensuring that personalization depends only on \emph{observed training structure} (no test leakage).

Each profile $p$ provides corresponding \emph{feature vectors} $f_k \in \mathbb{R}^{|\mathcal{U}_k|}$ that quantify preferences or relevance weights over the same attribute universes.  
The structure-gated bias is then:
\[
b_p(t) = \sum_{k=1}^{K} \alpha_k \, \langle w_k,\, g_k(t) \odot f_k \rangle,
\]
where $\odot$ denotes elementwise product, $w_k$ are small learnable weights, and $\alpha_k$ are scalar gates.  
This ensures that personalization affects only attributes the entity \emph{actually has}, providing both interpretability and rank variance.

%In matrix form:
%\[
%b_p = \sum_{k=1}^{K} \alpha_k \, G_k (w_k \odot f_k),
%\]
%which can be precomputed once per profile and broadcast-added to every query’s base scores at inference time.

\begin{figure*}[!h]
  \centering
  \includegraphics[width=0.8\linewidth]{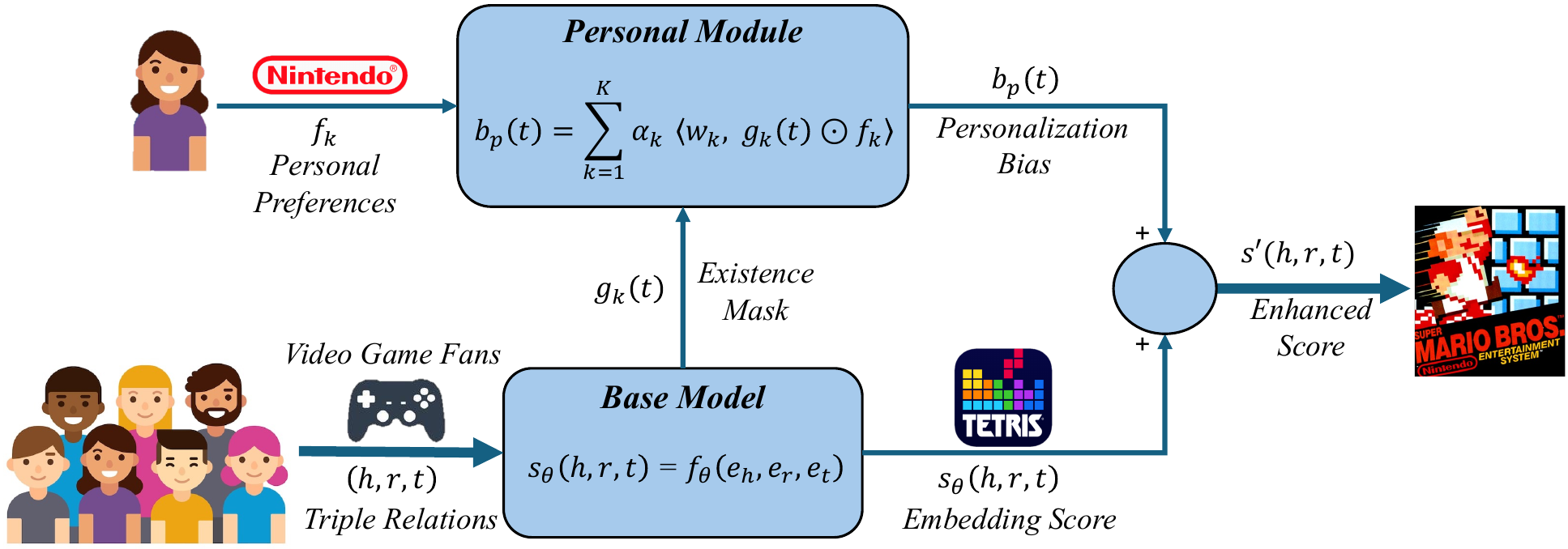}
\caption{Overview of the GatedBias framework. A frozen knowledge-graph model is refined by lightweight gates that combine structural and profile features to re-rank entities, enabling efficient, interpretable personalization without retraining.}
  \label{fig:sumblock}
\end{figure*}

\subsection{Profile Feature Construction}

The profile feature vectors $f_k$ encode how strongly a user values each attribute in $\mathcal{U}_k$.  
For interaction datasets (Amazon-Book, Last-FM), we construct $f_k$ via a three-stage process:

\begin{enumerate}
    \item \textbf{Individual preference extraction:}  
    For each user $u$, compute attribute frequencies within their interaction history $\mathcal{I}_u$:
    \[
    p_u(a) = \frac{1}{|\mathcal{I}_u|} \sum_{i \in \mathcal{I}_u} \mathbb{1}[a \in \text{attr}(i)].
    \]
    \item \textbf{Population aggregation:}  
    Aggregate across users: $w(a) = \sum_{u} p_u(a)$.
    \item \textbf{Feature normalization:}  
    Construct $f_k[j] = \text{clip}(\alpha \cdot w(a_j), 0, \tau)$ with scaling $\alpha$ and cap $\tau$.
\end{enumerate}

This yields dense, semantically meaningful profiles aligned to each relation group.  

%\paragraph{Feature Sparsity Mitigation.}
%Naïve universes $\mathcal{U}_k$ are extremely sparse ($<0.2\%$ nonzeros).  
%To ensure learnability, we filter $\mathcal{U}_k$ to include only entities appearing in user profiles and inject a small Gaussian baseline:
%\[
%f_k^{\text{final}} = f_k^{\text{semantic}} + \epsilon \nu, \quad \nu \sim \mathcal{N}(0, I), \; \epsilon = 0.02,
%\]
%achieving dense yet semantically faithful features and stable optimization.

\subsection{Training Objective and Optimization}

We optimize only $\{w_k, \alpha_k\}$ per profile while keeping $\theta$ frozen.  
For each batch $\mathcal{B}$ of positive and negative triples:
%\[
%\mathcal{L} = \frac{1}{|\mathcal{B}|} \sum_{(h,r,t) \in \mathcal{B}} 
%\max\big(0, 1 - [s'(h,r,t^+) - s'(h,r,t^-)]\big)
%+ \lambda_1 \sum_k \|w_k\|_1 + \lambda_2 \sum_k \|w_k\|_2^2,
%\]
\begin{align*}
\mathcal{L} &= \frac{1}{|\mathcal{B}|} \sum_{(h,r,t) \in \mathcal{B}} 
\max\big(0, 1 - [s'(h,r,t^+) - s'(h,r,t^-)]\big) \\
&\quad + \lambda_1 \sum_k \|w_k\|_1 + \lambda_2 \sum_k \|w_k\|_2^2,
\end{align*}
where $\lambda_1$ and $\lambda_2$ control sparsity and scale stability.  
Only a few hundred parameters are trained per profile ($\ll 0.1\%$ of backbone), making personalization ad-hoc, lightweight and reproducible. \textit{See Appendix for implementation details, hyperparameters, and training configuration.}

\subsection{Datasets and Profile Semantics}

We use the Amazon-Book~\cite{Ni2019Amazon} and Last-FM~\cite{HetRec2011LastFM} datasets as processed by KGAT~\cite{KGAT19}. Details are in Table~\ref{tab:datasets}.

\begin{table}[!h]
\centering
\setlength{\tabcolsep}{4pt}
\begin{tabular}{lrrrr}
\toprule
\textbf{Dataset} &
\textbf{Entities} &
\textbf{Relations} &
\textbf{KG Triples} &
\textbf{Users} \\
\midrule
Amazon-Book & 113{,}472 & 39 & 2{,}556{,}989 & 70{,}679 \\
Last-FM     & 58{,}266  & 9  & 464{,}567     & 23{,}566 \\
\bottomrule
\end{tabular}
\caption{Dataset statistics and experimental characteristics used for structure-gated personalization.}
\label{tab:datasets}
\end{table}

\paragraph{Amazon-Book (content preference vs.\ metadata).}
Entities represent books and associated attributes. We define two relation groups: \emph{content preference} (\textsc{genre}, \textsc{theme}, \textsc{subject}, \textsc{series}) and \emph{metadata} (\textsc{author}, \textsc{publisher}, \textsc{year}, \textsc{language}).

\paragraph{Last-FM (musical preference vs.\ metadata).}
Entities correspond to artists, tracks, and related descriptors. We define two relation groups: \emph{musical preference} (\textsc{genre}, \textsc{tag}, \textsc{style}, \textsc{mood}, \textsc{tempo}, \textsc{similar}, \textsc{influenced\_by}, \textsc{related}, \textsc{sound}) and \emph{metadata} (\textsc{label}, \textsc{year}, \textsc{country}, \textsc{language}, \textsc{format}, \textsc{release\_date}, \textsc{duration}, \textsc{album\_type}).

\textbf{Profile and gate construction.}
For both datasets, we partition relations into two semantic groups and construct binary gate vectors $g_A(t), g_B(t)$ from \emph{training} triples only (no test leakage), indicating which group-specific attributes connect to each tail entity $t$. User preference profiles $f_A, f_B$ are derived from interaction logs: we aggregate user-level preferences (e.g., tag/genre affinities, play-count patterns) into per-attribute scores, propagate these to attributes connected to preferred items in training, and obtain feature vectors by summing contributions over matched attributes. Both feature vectors are normalized. This setup tests whether personalization increases alignment with user preferences while preserving cohort ranking performance.

\textbf{Bias computation.} With learned weights $w_A,w_B$ and scalars $\alpha,\beta$, we precompute the tail bias
\[
b_p \;=\; \alpha\, G_A (w_A \odot f_A) \;+\; \beta\, G_B (w_B \odot f_B),
\]
on CPU in one pass.

\subsection{Evaluation Metrics}

We report both standard ranking metrics and personalization metrics. Our objective is to ensure that standard ranking performance remains stable with the introduction of personalization, while the personalization metrics capture individual-level improvements.

\paragraph{Standard Ranking Metrics.}
We follow the standard filtered evaluation protocol used in link prediction.  
For each query $(h,r,?)$, candidate tails $t$ already seen in training or validation are filtered out when ranking the true tail.  
We report Mean Reciprocal Rank (MRR) \cite{VoorheesTice2000}, Hits@$k$ \cite{Bordes2013TransE}, and Normalized Discounted Cumulative Gain (NDCG@$k$) \cite{JarvelinKekalainen2002} \textit{(definitions provided in Appendix)}.
These metrics evaluate overall link prediction performance and ensure cohort-level quality is preserved after personalization.

\paragraph{Personalization Metrics.}
To quantify profile-conditioned effects, we introduce and compute two complementary measures:

\textbf{Alignment@k.}  
Measures whether top-$k$ predictions favor entities whose attributes match the profile's preferences.
For each entity $t$, we compute group contributions $c_A(t) = \alpha_A \langle w_A, g_A(t) \odot f_A \rangle$ and $c_B(t) = \alpha_B \langle w_B, g_B(t) \odot f_B \rangle$ (vectorized as $c_A = \alpha_A G_A v_A$ and $c_B = \alpha_B G_B v_B$ where $v_k = w_k \odot f_k$).

We define the \emph{aligned set} as entities receiving strong, directional personalization:
\[
A = \left\{ t \,\Big|\, \max(c_A(t), c_B(t)) > 0, \; m(t) \geq \tau \right\},
\]
where $m(t) = |c_A(t) - c_B(t)|$ captures the strength of differential influence between the two feature groups, and threshold $\tau$ is the $P$-th percentile of margins among entities with at least one positive contribution ($P \in \{60, 70, 80\}$; higher $P$ = stricter selection).
Intuitively, aligned entities are those where (1) personalization provides a net positive push from at least one group, and (2) the two groups have strongly divergent effects, indicating clear directional preference.

%Alignment@k is the fraction of top-$k$ predictions that %fall in $A$.  
%We report $\Delta$Alignment@k = Alignment@k$_{\text{adapted}}$ - Alignment@k$_{\text{base}}$ with 95\% bootstrap CIs.
%A positive $\Delta$ indicates the adapter successfully promotes entities aligned with the profile's feature preferences.

%Measures whether top-$k$ predictions align with the profile’s preferred attribute groups.  
%We compute group contributions $c_A = \alpha_A G_A v_A$ and $c_B = \alpha_B G_B v_B$ where $v_k = w_k \odot f_k$.  
%Entities with positive margins $m(t)=|c_A(t)-c_B(t)|$ above a percentile threshold ($P\in\{60,70,80\}$) form the “aligned” set.  
%Alignment@k is the proportion of aligned entities in the top-$k$ predictions.  
%We report $\Delta$Alignment@k (adapted minus base) and 95\% bootstrap confidence intervals.  
%Statistical significance of $\Delta$Alignment@10 is tested using a two-sample t-test on seed-level means (equal variances, same number of seeds per condition).

\textbf{Counterfactual Responsiveness (CR).}  
To validate that personalization responds causally to features, we test whether entities currently benefiting from a feature group improve more when that group is boosted.
For each entity $t$, let $c_A(t) = \alpha_A \langle w_A, g_A(t) \odot f_A \rangle$ denote the contribution of group A to its bias (analogously for $c_B(t)$, with total bias $b_p(t) = c_A(t) + c_B(t)$).
Define $A^+ = \{t : c_A(t) > 0\}$ as entities whose scores are currently increased by group A features.

For each test query $(h, r, t^*)$, we measure the rank change of the ground-truth tail $t^*$ and categorize the query based on whether $t^* \in A^+$.
We apply a targeted perturbation by scaling group A features ($f_A \leftarrow (1+\epsilon)f_A$, e.g., $\epsilon=0.1$), recompute all ranks, and measure:
\[
\mathrm{CR}_A = \mathbb{E}[\Delta\text{rank} \mid t^* \in A^+] - \mathbb{E}[\Delta\text{rank} \mid t^* \notin A^+],
\]
where $\Delta\text{rank}$ is the change in rank for the ground-truth tail.
Since lower rank numbers are better (rank 1 is best), negative CR indicates correct responsiveness: test queries whose true answers are in $A^+$ show greater rank improvements (move to lower ranks) than others when we boost the features that already favor them.

%Evaluates causal sensitivity to targeted feature perturbations.  
%We scale one feature group ($f_A \leftarrow (1+\epsilon)f_A$) and recompute ranks.  
%CR measures the expected rank change between entities positively influenced by the perturbed group versus those outside it:
%\[
%\mathrm{CR}_A = \mathbb{E}[\Delta\text{rank} \mid t \in A^+] - \mathbb{E}[\Delta\text{rank} \mid t \notin A^+],
%\]
%where $A^+=\{t : c_A(t) > 0\}$.  
%Negative CR indicates beneficial responsiveness (true tails move up in rank).

\subsection{Baselines and Ablations}

We compare our structure-gated method to:

\textbf{Frozen Backbone.} The pretrained DistMult scorer with no adaptation: $s(h,r,t) = s_\theta(h,r,t)$. This establishes cohort-level performance without personalization.

\textbf{PatientNode.} A profile-agnostic ablation learns fixed entity biases with a lightweight MLP, $b(t)=\mathrm{MLP}_\phi(E_t)$ (same loss), ignoring $f_A,f_B$ and $g_A,g_B$—thus applying the same boost to all users (similar to always ranking pizza higher because it is a popular dish, regardless of whether the user is a meat-lover or vegan).

%An ablation that learns fixed, profile-agnostic entity biases via a lightweight MLP,
%$b(t)=\mathrm{MLP}_\phi(E_t)$, trained with the same ranking loss. This model boosts entities
%based only on embedding patterns, yielding the same bias across all profiles (similar to always ranking pizza higher because it is a popular dish, regardless of whether the user is a meat-lover or vegan). It tests whether personalization truly requires profile conditioning
%by lacking access to profile features $f_A,f_B$ and structural gates
%$g_A(t),g_B(t)$.

%An ablation that learns fixed, profile-agnostic entity biases. It trains a lightweight MLP to map frozen entity embeddings to scalar adjustments: $b(t) = \text{MLP}_\phi(E_t)$, optimized with the same ranking loss. This learns which entities should receive boosts based on their embedding patterns, but produces the same bias for all profiles (similar to always ranking pizza higher because it is a popular dish, regardless of whether the user is a meat-lover or vegan). Because it cannot condition on profile features $f_A, f_B$ or structural gates $g_A(t), g_B(t)$, this ablation tests whether profile-driven personalization is necessary. 

\section{Results and Discussion}
\begin{table*}[!htbp]
\centering
\setlength{\tabcolsep}{3pt}
\begin{tabular}{llccccc}
\toprule
\textbf{Method} & \textbf{Dataset} & \textbf{MRR} & \textbf{H@1} & \textbf{H@3} & \textbf{H@10} & \textbf{NDCG@10} \\
\midrule
{DistMult (Base)} 
    & Amazon-Book & 0.649{\tiny$\pm$0.002} & 0.622{\tiny$\pm$0.001} & 0.661{\tiny$\pm$0.002} & 0.675{\tiny$\pm$0.002} & 0.650{\tiny$\pm$0.001} \\
    & Last-FM     & 0.831{\tiny$\pm$0.002} & 0.760{\tiny$\pm$0.002} & 0.888{\tiny$\pm$0.001} & 0.949{\tiny$\pm$0.001} & 0.832{\tiny$\pm$0.001} \\
\midrule
{+ PatientNode}  
    & Amazon-Book & 0.653{\tiny$\pm$0.001} & 0.624{\tiny$\pm$0.002} & 0.662{\tiny$\pm$0.002} & 0.690{\tiny$\pm$0.001} & 0.656{\tiny$\pm$0.001} \\
    & Last-FM     & 0.828{\tiny$\pm$0.002} & 0.756{\tiny$\pm$0.002} & 0.884{\tiny$\pm$0.001} & 0.947{\tiny$\pm$0.001} & 0.829{\tiny$\pm$0.001} \\
\midrule
{+ GatedBias}    
    & Amazon-Book & 0.649{\tiny$\pm$0.001} & 0.622{\tiny$\pm$0.002} & 0.661{\tiny$\pm$0.001} & 0.675{\tiny$\pm$0.002} & 0.650{\tiny$\pm$0.001} \\
    & Last-FM     & 0.831{\tiny$\pm$0.002} & 0.759{\tiny$\pm$0.002} & 0.888{\tiny$\pm$0.001} & 0.950{\tiny$\pm$0.001} & 0.831{\tiny$\pm$0.001} \\
\midrule
\multicolumn{7}{r}{\textit{Parameter counts:} DistMult (400K), +PatientNode (400K+800), +GatedBias (400K+292)}\\
\bottomrule
\end{tabular}
\caption{Standard link prediction across datasets. GatedBias preserves performance with minimal overhead.}
\label{tab:link-pred}
\end{table*}

\begin{table}[!htbp]
\centering
\setlength{\tabcolsep}{1pt} % tighter horizontal spacing
\renewcommand{\arraystretch}{1.2}

\begin{tabular}{
  l
  S[table-format=1.3]
  S[table-format=1.3]
}
\toprule
\textbf{Personalization Metric} & \textbf{Amazon-Book} & \textbf{Last-FM} \\
\midrule
\multicolumn{3}{l}{\textbf{\textit{I. Alignment Effectiveness}} $\uparrow$} \\
\midrule
\hspace*{1.5em}Alignment@10 (Base)       & 0.064 {\tiny$\pm$ 0.003} & 0.831 {\tiny$\pm$ 0.002} \\
\hspace*{1.5em}Alignment@10 (Adapted)    & \textbf{0.073 {\tiny$\pm$ 0.003}} & \textbf{0.842 {\tiny$\pm$ 0.002}} \\
\hspace*{1.5em}$\Delta$ Alignment         & {\hspace*{-0.55em}+}0.009 {\tiny$\pm$ 0.001} & {\hspace*{-0.55em}+}0.011 {\tiny$\pm$ 0.001} \\
\hspace*{1.5em}$p$-value                  & 0.0210 & 0.0025 \\
\midrule
\multicolumn{3}{l}{\textbf{\textit{II. Counterfactual Responsiveness (CR)}} $\downarrow$} \\
\midrule
\hspace*{1.5em}Preference Gate CR          & {\hspace*{-0.55em}$-$}6.11 {\tiny$\pm$ 0.74} & {\hspace*{-0.8em}$-$}0.20 {\tiny$\pm$ 0.15} \\
\hspace*{1.5em}Metadata Gate CR            & {\hspace*{-0.55em}$-$}0.97 {\tiny$\pm$ 0.18} & {\hspace*{-0.8em}$-$}0.73 {\tiny$\pm$ 0.12} \\
\hspace*{1.5em}Preference: {\tiny\%} Improved & 28.5 {\tiny$\pm$ 0.7}{\tiny\%} & 3.0 {\tiny$\pm$ 0.5}{\tiny\%} \\
\hspace*{1.5em}Metadata: {\tiny\%} Improved   & 16.0 {\tiny$\pm$ 1.4}{\tiny\%} & 4.0 {\tiny$\pm$ 0.6}{\tiny\%} \\
\bottomrule
\end{tabular}

\caption{Personalization and Counterfactual Responsiveness for GatedBias.
Mean $\pm$ standard error over 5 runs. $\uparrow$: higher is better; $\downarrow$: lower is better.}
\label{tab:gated_bias_results}
\end{table}

\begin{table}[!h]
\centering
\setlength{\tabcolsep}{2.8pt} % tighter horizontal spacing
\renewcommand{\arraystretch}{1.1} % slightly tighter rows

\begin{tabular}{lccc} 
\toprule 
\textbf{Dataset} & 
\textbf{Real Features} & 
\textbf{Shuffled Features} & 
\textbf{Ratio} \\ & 
\textbf{$\Delta$ Alignment@10} & \textbf{$\Delta$ Alignment@10} & \\ 
\midrule 
Amazon-Book & {\hspace*{-0.35em}+}0.009 {\tiny$\pm$ 0.001} & {\hspace*{-0.35em}+}0.0004 {\tiny$\pm$ 0.0001} & $21.4\times$ \\ Last-FM & {\hspace*{-0.35em}+}0.001 {\tiny$\pm$ 0.0002} & {\hspace*{-0.35em}+}0.0001 {\tiny$\pm$ 0.00003} & $8.6\times$ \\ 
\bottomrule 
\end{tabular}

\caption{Placebo validation using frozen alignment masks with shuffled features. 
Near-zero placebo effects confirm the metric measures genuine personalization.}
\label{tab:placebo}
\end{table}

\textbf{Cohort Performance.} Table~\ref{tab:link-pred} evaluates whether adding personalization degrades the backbone's link prediction quality. GatedBias achieves link-prediction performance preservation on both datasets, while our ablation PatientNode shows inconsistent effects ($+0.004$ on Amazon-Book but $-0.003$ on Last-FM). Our approach requires only 292 additional parameters versus PatientNode's 800; establishing parameter-efficient personalization with no cohort-performance trade-off.

\textbf{Personalization Signal and Causal Validation.} Table~\ref{tab:gated_bias_results} shows that our mechanism successfully reorders candidates toward profile-aligned entities and \emph{validates} the causal pathway. Amazon-Book demonstrates strong, validated personalization: alignment increases $+14\%$ relative ($+0.9$pp absolute, from $6.4\%$ to $7.3\%$, $p=0.021$). Counterfactual perturbations confirm causality by boosting preference features improves in-group entities by \emph{$CR=-6.11$ rank positions} (lower is better; more negative indicates stronger improvement), with $28.5\%$ of queries showing measurable improvement. On Amazon-Book, preference gates (\textsc{genre}, \textsc{theme}, etc.) dominate metadata gates ($CR=-6.11$ vs.\ $-0.97$; $28.5\%$ vs.\ $16.0\%$ improved).

Last-FM shows weaker but significant effects: $+1.3\%$ relative alignment lift ($+1.1$pp, $p=0.0025$) and minimal causal responsiveness. Using the same convention (lower is better), metadata gates \emph{outperform} preference gates on Last-FM: $CR=-0.73$ with $4\%$ improved versus preference $CR=-0.20$ with $3\%$ improved, inverting the Amazon-Book pattern. The stark CR gap between datasets (about $6.11$ vs.\ $0.20$ in magnitude) highlights domain constraints. Last-FM's $\sim 83\%$ baseline alignment indicates the frozen backbone \textit{already} captures user–item affinity well which leaves minimal headroom for personalization. Incremental gains (e.g., from $83\%$ to $84\%$) face ceiling effects unless signals are extremely precise. Moreover, music choices are inherently noisy (mood, context, serendipity), so collaborative signals already capture much of the actionable variance; explicit preference features add little. By contrast, product catalogs are sparser and more structured, so semantic attributes (e.g., \textsc{genre}, \textsc{influenced\_by}) provide clearer leverage for personalization.

\textbf{Placebo Validation.} Table~\ref{tab:placebo} provides the critical validity check: if our alignment metric genuinely measures feature-grounded personalization, shuffling profile features while freezing the alignment mask should collapse effects to near zero. With randomly shuffled features, alignment gains drop dramatically. Real features produce $21.4\times$ stronger effects than noise on Amazon-Book and $8.6\times$ on Last-FM.

%%%%%%%%%%%%%%%%%%%%%%%%%%%%%%%%%%%%
\section{Conclusion}

Our method achieves statistically significant personalization effects while maintaining cohort performance. Effect sizes are modest in absolute terms, but placebo validation confirms they represent genuine feature-driven signal rather than artifacts, with real features yielding order-of-magnitude stronger effects than random noise across both datasets. Results reveal clear domain dependencies that inform deployment strategies. Future work should explore better feature engineering and extensions to head/relation conditioning to boost effect sizes.

%\textbf{Implications and Design Principles.} Our results establish four key contributions for lightweight KG personalization: (1) \textbf{Zero-cost adaptation}: perfect cohort preservation with 292 parameters enables risk-free deployment alongside frozen backbones, with the PatientNode ablation demonstrating that profile-independent approaches sacrifice this stability. (2) \textbf{Validated personalization}: statistically significant effects ($p<0.05$) with 8–21× placebo ratios confirm genuine reordering beyond what fixed entity biases can achieve. (3) \textbf{Domain characterization}: the method is most effective in sparse, uncertain domains (low base alignment, strong feature-outcome relationships) and less suitable for dense, mature systems. (4) \textbf{Interpretable mechanism}: structure-gated design makes explicit which entity attributes drive personalization (e.g., content features 6× more effective than metadata), unlike embedding-based approaches that learn opaque entity adjustments.  Future work should explore stronger feature engineering, dynamic gate selection, and extensions to head/relation conditioning to increase effect magnitudes while preserving the zero-cost guarantee.

%\section{Acknowledgments}
%\bigskip
%\noindent asd

\bibliography{aaai2026}
%\newpage

\appendix
\newpage
\section{Appendix}

\subsection{Implementation Details}

Experiments use pretrained DistMult embeddings as frozen backbones.  
Training on top of frozen backbone employs batch size $4096$, learning rate $10^{-3}$, and five epochs.  
Evaluation uses mixed precision on a single NVIDIA V100 GPU.  
We repeat all runs with three random seeds and report mean $\pm$ standard deviation.  
Each method (Base, NoGates, GatedBias) is evaluated under identical conditions and random seeds for fair comparison.  
Results include MRR, Hits@$k$, NDCG@$k$, $\Delta$Alignment@k, and CR with $\epsilon = 0.1$. For normalization we used scaling $\alpha = 0.1$ and cap $\tau = 0.5$.

\subsection{Definition of Standard Ranking Metrics} \label{def_metrics}

\[
\text{MRR} = \frac{1}{|Q|} \sum_{q \in Q} \frac{1}{\text{rank}(q)}, \quad
\text{Hits@}k = \frac{1}{|Q|} \sum_{q \in Q} \mathbb{1}[\text{rank}(q) \le k],
\]
\[
\text{NDCG@}k = \frac{1}{|Q|} \sum_{q \in Q} \frac{\text{DCG@}k(q)}{\text{IDCG@}k(q)}.
\]

\end{document}